\def\year{2018}\relax
\documentclass[letterpaper]{article} 
\usepackage{aaai18}  
\usepackage{times}  
\usepackage{helvet}  
\usepackage{courier}  
\usepackage{url}  
\usepackage{graphicx}  
\begin{document}

\title{Traffic Density Estimation using a Convolutional Neural Network \\ \em Machine Learning Project - National University of Singapore}
\author{Julian Nubert$^1$, Nicholas Giai Truong$^2$, Abel Lim$^3$, Herbert Ilhan Tanujaya$^3$, Leah Lim$^3$, Mai Anh Vu$^3$ \\ April 2018 \\ \small $^1$\em ETH Zurich, $^2$University of Southern California, $^3$National University of Singapore}
\nocopyright 
\maketitle

\begin{abstract}
  The goal of this project is to introduce and present a machine learning application that aims to improve the quality of life of people in Singapore. In particular, we investigate the use of machine learning solutions to tackle the problem of traffic congestion in Singapore.

  In layman's terms, we seek to make Singapore (or any other city) a smoother place. To accomplish this aim, we present an end-to-end system comprising of 
  \begin{enumerate}
  \item A traffic density estimation algorithm at traffic lights/junctions and
  \item a suitable traffic signal control algorithms that make use of the density information for 			better traffic control.
  \end{enumerate}

  Traffic density estimation can be obtained from traffic junction images using various machine learning techniques (combined with CV tools). After research into various advanced machine learning methods, we decided on convolutional neural networks (\textit{CNNs}). We conducted experiments on our algorithms, using the publicly available traffic camera dataset published by the Land Transport Authority (LTA) to demonstrate the feasibility of this approach. With these traffic density estimates, different traffic algorithms can be applied to minimize congestion at traffic junctions in general.
\end{abstract}

\section{Real World Application Scenario}
\label{section:application}

In this report, we present and discuss a potential application which estimates the traffic density at traffic lights/junctions using public cameras to adapt the traffic lights accordingly to get the best result.

\subsection{Justification}

Even if traffic is flowing very slowly, streets could handle the traffic flow much more efficiently; this means either more traffic at the same time or the same traffic in a shorter time. The key for this is that all cars have to move at constant speed without much braking and accelerating. Therefore, an intelligent traffic system could detect the amount of cars at every position, estimate the velocity of the cars in a later stage, and ultimately adapt the traffic lights accordingly to get the optimal outcome. As described in \cite{topdown} a \emph{top down traffic control}, which is used in general, is completely centralized and its control schemes are developed off-line.
\begin{quote}
  \emph{The problem of this top-down control based on specific scenarios triggered according to some patterns is that they hardly fits well in practice.} \cite{topdown}
\end{quote}
Just think about extraordinary situations such as changing weather, accidents or other unplanned traffic fluctuations. Efficiency can be gained by locally adapting the traffic lights, considering the local traffic situation.

\subsection{Significance}

We identified several reasons why there is a need for this application in Singapore. For us, the crucial points are the following:
\begin{itemize}
  \item This application will help everyone who moves around Singapore frequently, so it is universally beneficial.
  \item It reduces the time and cost of traffic congestion.
  \item The higher efficiency in traffic and less traffic jams also have a positive impact on the climate (by reducing greenhouse gases emissions such as CO$_2$).
  \item It is useful for future integration with autonomous vehicle technology since it paves the way for an efficient "fleet management".
  \item The required infrastructure (cameras on top of traffic lights) is publicly available for Singapore and could be easily used.
\end{itemize}

\subsection{Requirements}

For our system to be employable some certain requirements must be fulfilled. In the following we show the necessities for our \emph{intelligent traffic control}:
\begin{enumerate}
  \item Real Time: Receiving a camera image must lead to an instantaneous estimation and to the needed Traffic light adaption.
  \item Fail Safeness: Since a wrongly working traffic light system is highly dangerous it must be absolutely \textit{failsafe}.
  \item Superior Rules: It is still necessary to introduce some \textit{rules} to avoid wrongdoing, e.g. to avoid starving of cars.
  \item Work under different conditions: Our software must be \textit{versatile} and should work in \textit{different situations} (changing lighting, weather and traffic conditions) as well as at \textit{different places}.
  \item Streaming Data: We are constantly receiving data by the cameras. Therefore, we must be able to perform \emph{Stream Processing} (incrementally).
\end{enumerate}

\subsection{Human-Application Interface}

The \textit{first and designated interface} between the application and the involved human drivers is quite obvious. The system just gives the same outputs as a normal traffic light. People then just follow this regulation as they did previously. Therefore, the system helps the humans in this case, and they don't have to pay attention to any additional signs.
The \textit{second part of the interface} involves pedestrians. What happens if people want to cross the road? If there are provided pedestrian lights, we then simply add an additional input to our pipeline. If not, there are two possibilities: Try to perceive them using the camera as well and include them in our traffic decisions or to just ignore them. Both have valid reasons, and the decision depends on individual circumstance (e.g. compare a motorway to a play street).

For the latter case we therefore need to expand our decision policy.

\subsection{Ethical Implications}

We think that our application is not too critical in this respect, which is also a reason for us to pick this specifically.

The application does not displace jobs because it simply improves existing traffic algorithms. Camera images are already available publicly, and training on them presents no privacy violation.

The potential concern with this application is the possibility of exploitation for malicious intent. E.g. consider a scenario where a party wants to use this application commercially and to privilege some cars who have paid large amounts of money, leading to inequity. Hence, it is likely better to let the authorities be in charge of this application.

Also, not being vulnerable to hacking attacks or expanding the service to more critical activities would be one of our important objectives.

\section{Algorithmic Structure}

We divide our application in two main topics:
\begin{enumerate}
  \item Traffic Density Estimation and
  \item Decision Making based on the Estimation.
\end{enumerate}
The \textit{first part} receives the live camera image of every lane facing towards the traffic junction. Using this information, it then deals with determining the traffic density on each of the lanes.

Using this information, the task of \textit{the second part} is then to set up the \emph{optimal} traffic light state considering also all of the requirements specified in the \emph{Requirements} section.

In the next sections we will present both parts; however, our main focus will be on the first one. For this one, there's no way of getting around machine learning algorithms. Therefore, we present our own pipeline, show how we approached this problem and will discuss how the results differ from our expectations. For the second part, we will discuss existing approaches and their suitability.

\section{Estimation}

\subsection{Approach}
As shortly described above, in the practical application the estimation part would receive a live image stream of every lane intersecting the junction. In our case, it was very difficult to find an appropriate training set in general and specifically for Singapore. Therefore we decided to use the live camera data from Singapore (LTA)\footnote{\label{SingDatasetFootnote}https://data.gov.sg/dataset/traffic-images}.

\begin{figure}[ht]
  \includegraphics[width=1.0\linewidth]{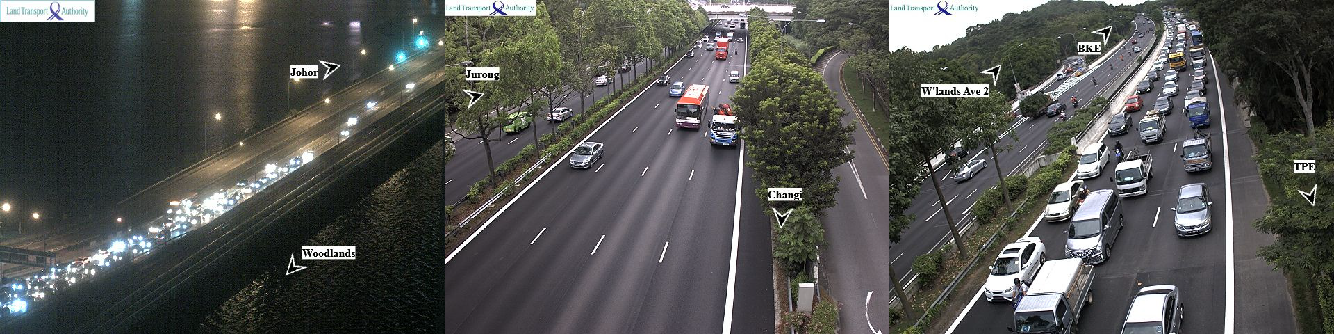}
  \caption{Image from Camera1/Camera2/Camera3 at Night/Day/Day with High/Low/Traffic Jam Density}
  \label{fig:CamImageSample}
\end{figure}

We wrote a script to download images from all of the cameras over a weekend and selected three cameras which seemed to be the most suitable for our use case; here we chose those that have varying traffic density over the days and contains a clearly visible road (i.e.\ unobstructed by trees, etc). We then used these images and randomly partitioned the dataset into 90\% for \textit{training} and 10\% for \textit{validation}. We decided to use images taken during both day time and night time. You can find 3 sample images of the three different situations with different lighting and density conditions in figure~\ref{fig:CamImageSample}. Overall we had \textbf{4582 images} available.
We decided to define 5 classifiers to categorize the images. They can be found in Table~\ref{table:classifiers}. We counted motorcycles as half cars.
\begin{table}[ht]
  \centering
  \begin{tabular*}{1.0\textwidth}{llr}
    \textbf{Classifier} & \textbf{Meaning} & \textbf{Definition} \\
    Empty & Almost empty street & 0-8 Cars \\
    Low & Only a few cars & 9-20 Cars \\
    Medium & Slightly filled street & $<$ 50 cars \\
    High & Filled Street or Blocked Lane & $<$ 100 cars \\
    Traffic Jam & Traffic almost not moving & $>$ 100 cars \\
  \end{tabular*}
  \caption{Definition of Traffic Density}
  \label{table:classifiers}
\end{table}

\subsection{Model}

Amongst all the proposed advanced machine learning topics, \emph{Convolutional Neural Network} was the most suitable approach for us.

In addition to the choice of this model there are also many other possibilities available. We thought about the following possibilities:
\begin{itemize}
  \item Feeding the machine learning pipeline with the raw image or with some extracted features (SIFT, SURF, etc.).
  \item Preprocess the image (cut off unimportant parts or not).
  \item Grayscale or colored image.
  \item Resolution of the image.
  \item Structure of the underlying Neural Network (activation functions, number of layers, etc.), see section \emph{ML Model in our Case} for more specific analysis.
\end{itemize}

\subsection{Alternative Model}

For us, the most suitable of the other alternative models would have been the \emph{Recurrent Neural Network}. Due to its structure in which connections form a directed graph along a sequence, for \emph{RNNs} it's possible to use their internal state as a memory. This allows processing sequences of inputs\footnote{https://en.wikipedia.org/wiki/Recurrent\_neural\_network}.

This could be suitable to even input multiple sequential images into the pipeline and hence be able to estimate the velocity.

However, we did not attempt this because the Singapore live dataset (see footnote \ref{SingDatasetFootnote}) were at 20 seconds interval, which we considered to be too long for RNNs to be used effectively. Thus, we chose to focus on traffic density estimation in this project.

\subsection{Technical Details and Insights}

\emph{Convolutional neural networks}, or \emph{CNNs} in short, are a special form of Neural Networks. They are especially well suited for the processing of inputs which have a grid-like topology. According to \cite{Goodfellow-et-al-2016}
\begin{quote}
  they have been tremendously successful in practice applications.
\end{quote}
The same book defines \emph{CNNs} simply by the following quote:
\begin{quote}
  Convolutional networks are simply neural networks that use convolution in place of general matrix multiplication in at least one of their layers.
\end{quote}
In this discussion we will mainly focus on the distinctions and the advantages this topology induces in comparison to \emph{general neural networks}.
\subsubsection{Convolution}
Normally a mathematical convolution is denoted as:
\begin{equation}
  \label{equ:conv}
  s(t) = (x \ast w)(t)
\end{equation}
In the case of CNNs, the first argument in equation~\ref{equ:conv} is most of the time referred to as the \emph{input}, whereas the second argument as the \emph{kernel}. The output is sometimes called \emph{feature map}.

\subsubsection{Motivation}

There are three important ideas in existence why convolution can help improve a machine learning system, in our case here \emph{CNNs} compared to the typical \emph{neural networks} (typical \emph{NNs}):
\begin{enumerate}
  \item Sparse Interactions,
  \item Parameter Sharing and
  \item Equivariant Representations.
\end{enumerate}
Moreover, the convolution provides a good possibility to process variable sized inputs.

\subsubsection{Details}

\begin{figure}[ht]
  \includegraphics[width=1.0\linewidth]{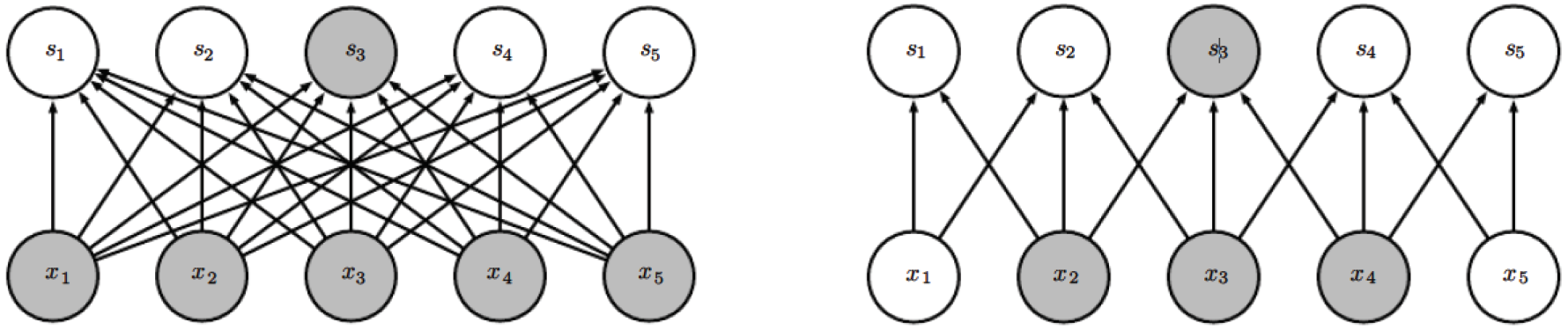}
  \caption{Units affecting the output unit; \textbf{Left:} Formed by \emph{normal matrix multiplication}; \textbf{Right:} Formed by \emph{convolution}, see \cite{Goodfellow-et-al-2016}}
  \label{fig:NNsComparison}
\end{figure}

Compared to typical \emph{NNs}, where every input unit interacts with every output unit, \emph{CNNs} use \textbf{sparse interactions}. The reason for this is that the \emph{kernel} is chosen to be smaller than the input. This means we need to store fewer parameters which reduces requirements in both memory and calculations. According to \cite{Goodfellow-et-al-2016} the improvements in efficiency are usually quite large. The difference between conventional \emph{NNs} and \emph{CNNs} is illustrated in figure~\ref{fig:NNsComparison}.

\begin{figure}[ht]
  \centering
  \includegraphics[width=0.35\linewidth]{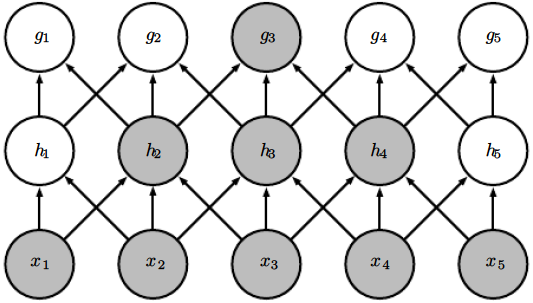}
  \caption{Illustration of Deeper Interaction of Neurons, see \cite{Goodfellow-et-al-2016}}
  \label{fig:CNNsDeepInteraction}
\end{figure}

It is important to mention that neurons in deeper layers may still indirectly interact with a larger portion of the input. This enables the successful consideration of complicated interactions between the \emph{simple building blocks} and hence the detection of more complicated structures in the input. For clarification refer to figure~\ref{fig:CNNsDeepInteraction}.

\textbf{Parameter Sharing} describes the usage of the same parameter for multiple model functions. It is also often denoted as \emph{tied weights} since the weight values applied to one input value is simply tied to a weight value applied somewhere else. In more detail, this means that \emph{each kernel is used at every position of the input} (except maybe the boundaries). Therefore we learn only one set of parameters. This further reduces the storage requirements. According to \cite{Goodfellow-et-al-2016},
\begin{quote}
  Convolution is thus dramatically more efficient than dense matrix multiplication in terms of the memory requirements and statistical efficiency.
\end{quote}
By result of \emph{parameter sharing}, we observe another important property, the \textbf{equivariance} to translation. This property ensures that pixel shifts in an image does not affect the output. If we define $I'(x,y) = I(x-\Delta x,y)$ as the shifted image, now it makes no difference if we apply the convolution to the shifted image $I'$ or if we apply the convolution to the original image $I$ and then shift it.

\subsubsection{Pooling}

The third stage of each layer, after performing the convolutions and applying the activation functions, is often called \textbf{pooling}. For this, a \emph{pooling function} is used which replaces the output at a certain location with a summary statistic of nearby outputs\footnote{Typically, pooling is done by extracting the \emph{maximum value}, the \emph{average}, the \emph{$l_2$ norm} of the rectangular neighborhood or any other \emph{weighted average} (e.g. based on the distance from center)}. It always helps to make the output almost invariant to small translations, and for many tasks it is just essential to make the network applicable to inputs of varying size.

For describing the mentioned properties we used the book of \cite{Goodfellow-et-al-2016}, which we highly recommend.

\subsection{Advantages and Exploitation of Model}

We model the problem as a task of classifying images into 5 traffic density classes. Based on the properties of CNNs, they are suitable for this. This is especially true given the definition from above that \emph{CNNs} are
\begin{quote}
  especially well suited for the processing of inputs which have known grid-like topology. \cite{Goodfellow-et-al-2016}
\end{quote}

Also the size of the \emph{Kernel} for the convolution is a big advantage of using \emph{CNNs}. An image might have thousands or millions of pixels, but we can detect small and meaningful features (e.g. edges, corners) with kernels that are only consisting out of tens or hundreds of pixels.

Also the \emph{parameter sharing} is a very nice property in our case since it just reduces the amount of parameters significantly which would be a lot having a whole image as an input and using typical \emph{NNs}.

The translational \emph{equivariance} can also be very helpful due to the general reasons mentioned above.

However the \emph{pooling} still has no negative effects since in our case the exact location of the crucial structures (for detecting the cars) is not fixed, since cars are moving anyway. So one of the typical big disadvantages of pooling, perturbing the performance in situations in which the very exact location is important, is not an issue.

\subsection{ML Model in our Case}
\label{SuitableForUs}

State-of-the-art models have been empirically demonstrated to have good performance on general image classification tasks \cite{szegedy2016rethinking}. However, training these large models from scratch on our dataset is slow and prone to overfitting.
Instead, we can explore the use of transfer learning~\cite{yosinski2014transferable}. We make use of pre-trained weights from an \emph{InceptionV3} model by removing its penultimate layer and training a new softmax layer on top of it to produce predictions for our task. This allows us to make use of \emph{InceptionV3} to extract general image features for us and to train new models very quickly since we only have to train the additional layers.

\section{Decision Making}

With the traffic estimation approaches designed, we demonstrate that there exist basic approaches for using these estimates to solve the traffic algorithm problem.

\subsection{Possible Approaches}
\subsubsection{Reinforcement Learning} One of those possibilities was shown by \cite{DBLP:journals/corr/GaoSLIS17} where they proposed a \emph{deep reinforcement learning algorithm} which extracts all useful \emph{machine crafted} features from raw real-time traffic data. The goal was to learn the optimal policy to adapt the traffic lights. Impressively, they were able to reduce the vehicle delay by up to 47\% compared to the well known \emph{longest queue first algorithm} and even by up to 86\% compared to \emph{fixed time control}. The key behind this approach is the formulation of the traffic signal control problem as a reinforcement learning problem. In this case the goal of the \emph{agent} is to reduce the \textbf{vehicle staying time} in the long run. The \emph{reward} for the agent is given at each time step for choosing actions that decrease the time of vehicles staying at the intersection.
\subsubsection{Genetic Algorithms} Another possibility, which is admittedly not really state of the art, is using the \emph{Genetic Algorithms} as proposed by \cite{Singh}. This paper is presenting a strategy which is giving appropriate \emph{green time extensions} to minimize a fitness function. In this case the fitness function is consisting of a linear combination of performance indexes of all four lanes used in this example. This approach reaches in this paper a performance increase of 21.9\% which is not as good as the \emph{reinforcement policy} from last chapter.

\subsection{Training and Ways of Finding the Solution}
With the traffic algorithm readily available, we would also require proper simulation environments.
\subsubsection{Aimsun} The first simulation we tried was the \emph{aimsun next} traffic modeling software \footnote{https://www.aimsun.com/aimsun-next/}. The tool allows whole big cities can be imported and simulated which we considered to be too massive in scale for our use-case.
\subsubsection{SUMO} 
\begin{figure}[ht]
  \centering
  \includegraphics[width=0.35\linewidth]{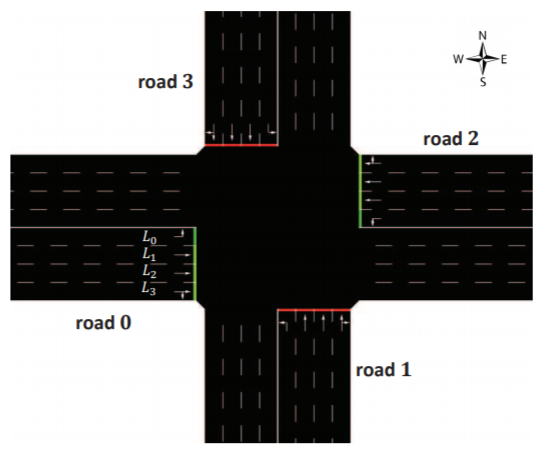}
  \caption{GUI of the SUMO simulation \cite{DBLP:journals/corr/GaoSLIS17}}
  \label{fig:simulation}
\end{figure}

Another better alternative is the popular open source simulator \emph{Simulation of Urban MObility (SUMO)}\footnote{http://sumo.dlr.de/userdoc/}. SUMO is an easy program with python API, can be configured using simple \emph{xml}-files, controlled using the terminal. This allows not only to verify the results but to also actively use this simulation during the training, e.g. for the case when a \textbf{Reinforcement Learning} approach is used as presented in \cite{DBLP:journals/corr/GaoSLIS17}. It's also possible to visualize the results in a GUI, an example image can be seen in figure~\ref{fig:simulation}.

\section{Tests and Experiments}

\subsection{Dataset}
We use self-labeled traffic images from 3 cameras of different junctions and angles as seen in figure~\ref{fig:CamImageSample}. Each image is labeled with a density level from empty to traffic jam.

\subsection{Setup}
We ran experiments using models built upon the \emph{Keras} library with \emph{TensorFlow} backend. \emph{Keras} provides us with an easy \emph{API} for building deep learning models which allowed us to focus more on the experiments.

\subsection{Approaches}
Traffic density estimation can be modeled as a \emph{multi-class} classification problem and be solved by \emph{CNN} classifiers. We identified the following approaches for making use of \emph{CNNs} and investigated their effectiveness for this problem:
\begin{enumerate}
  \item Basic \emph{CNN} and
  \item Transfer Learning on \emph{InceptionV3}.
\end{enumerate}

\subsection{Result}
We trained the two classifiers on the dataset and evaluated them on a few metrics\footnote{All metrics were measured by training a model on the \emph{training set} and performing evaluation on a separate \emph{validation set}. We take the average of all results over 10 runs}: \emph{accuracies}, \emph{f1 scores}\footnote{We extend \emph{F1 scores} to the \emph{multi-class} scenario by taking the average of all independently computed \emph{F1 scores} for each class. Each \emph{F1 score} is taken to be the harmonic mean of precision and recall for that class} and \emph{top 2 accuracies}\footnote{\emph{Top 2 accuracy} refers to the frequency that an example was correctly labeled by the \emph{rank 1-2} predictions}.
All results shown in Table~\ref{table:experiment results 1} are evaluated on a cross-validated classifier with hyperparameters selected based on \emph{accuracy}. The cross-validation was performed using a simple grid-search.
\begin{table}[ht]
  \centering
  \begin{tabular}{|c|c|c|c|}
    \hline
    \textbf{Classifier} & \textbf{Accuracy} & \textbf{F1} & \textbf{Top 2 Accuracy} \\ \hline
    Basic CNN & 71.35 & 71.26 & 93.23 \\ \hline
    Transfer Learning & 66.38 & 59.21 & 88.43 \\ \hline
  \end{tabular}
  \caption{Classifier accuracy results}
  \label{table:experiment results 1}
\end{table}
\begin{table}[ht]
  \centering
  \begin{tabular}{|c|c|c|}
    \hline
    \textbf{Classifier} & \textbf{Training Time} & \textbf{Training Time} \\
    \textbf{} &\textbf{/min} & \textbf{(With GPU)/min}\\ \hline
    Basic CNN & 40.8 & 1.1 \\ \hline
    Transfer Learning & 1.2 & 0.65\\ \hline
  \end{tabular}
  \caption{Time Efficiency Results}
  \label{table:Time Efficiency Results}
\end{table}
In Table~\ref{table:Time Efficiency Results} you can find the time the training took us with and without GPUs.

A simple \emph{CNN} provided an overall better performance than transfer learning on \emph{InceptionV3}. This is likely because we froze the entire \emph{InceptionV3}'s weights and higher-level features from the larger dataset cannot be transferred to our dataset \cite{yosinski2014transferable}. In the future, we could explore freezing the bottom $k$ layers only.

Although using a \emph{CNN} provides better overall accuracy than transfer learning, it take a significantly longer time to train without a GPU.\footnote{All time measurements are for 50 epochs of training with no extra preprocessing. GPU measurements were conducted on an \emph{Nvidia GTX1080Ti}} Therefore, transfer learning is a \textbf{viable approach if computation power is limited}. While CNN is the preferred approach when GPUs are available.

\subsection{Other Possibilities Tested}

\begin{table}[ht]
  \centering
  \begin{tabular}{|c|c|}
    \hline
    \textbf{Class} & \textbf{Count} \\ \hline
    Empty & 1679 \\ \hline
    Low & 1306 \\ \hline
    Medium & 556 \\ \hline
    High & 554 \\ \hline
    Traffic Jam & 488 \\ \hline
  \end{tabular}
  \caption{Number of traffic images per class}
  \label{table:experiment results}
\end{table}
\subsubsection{Uneven Distribution:} Due to uneven distributions of classes in our self-labeled dataset (see Table~\ref{table:experiment results}), we also explore the following measures to handle class imbalance (CI) and compare them through experiments:
\begin{enumerate}
  \item \textbf{Ratio-weighted losses:} We scale the \emph{cross entropy} losses contributed each example according to their class ratios using the following formulation:
    \begin{equation}
      \alpha_c = median\_count / count_c
    \end{equation}
    This increases the cost of misclassification of a minority class, forcing the learner to prioritize the correct classification of minority classes \cite{DBLP:journals/corr/EigenF14}.
  \item \textbf{Real-time data augmentation:} By performing basic image transformations on existing data, we are able to generate new examples on-the-fly for training to increase the variety of examples seen by the classifier. As it can be seen in Table~\ref{table:experiment results 2}, this leads to a better accuracy for the minority classes as we are able to obtain more training examples for them \cite{Wong2016UnderstandingDA}.
\end{enumerate}
\begin{table}[ht]
  \small
  \centering
  \begin{tabular}{|c|c|c|c|}
    \hline
    \textbf{Method} & \textbf{Accuracy} & \textbf{F1} & \textbf{Top 2 Accuracy} \\ \hline
    Basic CNN with & & & \\
    class imbalance & 73.0 & 80.52 & 93.56 \\
    measures applied & &  &  \\ \hline
  Basic CNN & 71.35 & 71.26 & 93.23 \\ \hline  \end{tabular}
  \caption{Class imbalance results}
  \label{table:experiment results 2}
\end{table}
With CI measures, accuracy increased slightly but f1 scores\footnote{F1 score is a better predictor of performance than accuracy for class imbalance scenarios since it accounts for precision and recall} increased significantly. Therefore, CI measures have shown significant improvements. Notably, the top 2 accuracy is about the same which indicates that class imbalance does not affect the top 2 predictions.
\subsubsection{Image Preprocessing:} Since traffic images consists of 2 opposite traffic lanes, we also propose the use of image masking\footnote{A visualization of masked preprocessed image can be found here: https://youtu.be/KA4SbJVX0mc} to remove parts of the images that are not in the interested traffic lane.
\begin{table}[ht]
  \small
  \centering
  \begin{tabular}{|c|c|c|c|}
    \hline
    \textbf{Method} & \textbf{Accuracy} & \textbf{F1} & \textbf{Top 2 Accuracy} \\ \hline
    Basic CNN with & &  &  \\
    CI measures & 74.3 & 81.3 & 94 \\
    and masking & &  &  \\ \hline
  \end{tabular}
  \caption{Image masking results}
  \label{table:experiment results 3}
\end{table}

\begin{figure}[ht]
  \includegraphics[width=1.0\linewidth]{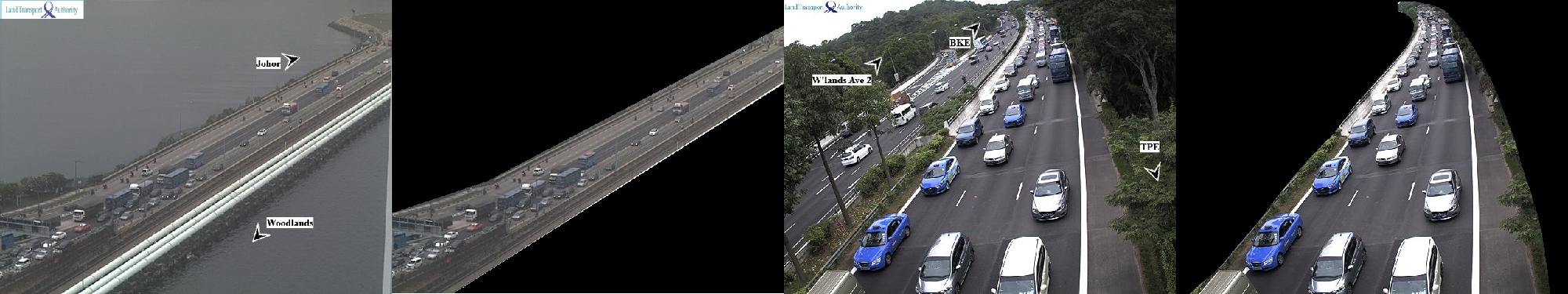}
  \caption{Original and Corresponding Masked Images}
  \label{fig:masking}
\end{figure}

As seen in Table~\ref{table:experiment results 3} the use of masking provided a 1-2\% increase in accuracy which is not very significant. This shows that the CNN model was able to identify the non-relevant parts of the image even without masking. An example of the masking can be seen in Figure~\ref{fig:masking}.

\subsection{Other Tools, Online Resources}
We made use of a data-labelling tool, \emph{Labelbox}\footnote{https://www.labelbox.io/}. It provides a user-friendly web interface for us to collaboratively label the entire dataset.

For the experiments and implementation, besides \emph{Keras}, we used \emph{OpenCV} for general image processing (e.g. masking the non-relevant parts of the images and resizing the image). \emph{matplotlib} was also used for general data visualization.

Our most important online resource was the Singaporean live camera dataset (see footnote~\ref{SingDatasetFootnote}). We wrote all scripts for downloading, processing and classifying these images by ourselves.

\section{Organization}
Manpower for the project was managed by assigning each person to a task that they were suited for. In the brainstorming phase, everyone was given time to develop their own ideas and to choose among all the ideas by a majority vote decision.

Moreover, unpleasant tasks (such as e.g. the \emph{labeling} of the data) were also divided equally among the members.

\section{Conclusion}

\subsection{Reaching Requirements}

Looking at the final results in Table~\ref{table:experiment results 3} we are happy to reach these numbers. It was quite a long way to get to this point with only having 4582 images available. An \textbf{accuracy} of 74.3\% and an \textbf{F1 score} of 81.3\% are already quite satisfying and in practice when having a \textit{frame rate} of a few images a second, the average classification (like a "low pass filter") over a certain time interval will very likely produce good results. Also for us the 94\% \textbf{Top 2 accuracy} is very significant because we labeled the images intuitively, hence the tendency is almost as important as the exact classification. E.g. if a \emph{low density} is classified as \emph{empty} it's still a good insight. The high \emph{Top 2 accuracy} just approves this.

Of course before bringing the application to the markets, further practice tests would be required. But we are very confident that its potential is high while it's also not too hard to guarantee the general requirements for this case (e.g. real time behavior).

\subsection{Future Improvements}

A big improvement would be to just use a larger training set which only hardly would have been possible for us because of the limited time.

For bringing up the \emph{Accuracy} as well as the \emph{F1 scores} it would also be helpful to do the labeling more precisely, i.e. really count the number of cars. But as mentioned in the last chapter, also classifying the right tendency can be very helpful for our application.

\fontsize{9.0pt}{10.0pt}
\selectfont
\bibliography{citation}
\bibliographystyle{aaai}
\end{document}